\documentclass[10pt,twocolumn,letterpaper]{article}

\usepackage{cvpr} % uncomment this for the final submission
\usepackage{times}
\usepackage{epsfig}
\usepackage{graphicx}
\usepackage{amsmath}
\usepackage{amssymb}
\usepackage{booktabs}
\usepackage{multirow}

\usepackage[pagebackref=true,breaklinks=true,letterpaper=true,colorlinks,bookmarks=false]{hyperref}

\begin{document}

%%%%%%%%% TITLE

%\title{DeepFake on Realistic Scenarios: Semantic Mismatch as a Novel Challenge}
\title{Are DeepFakes Realistic Enough? Exploring Semantic Mismatch as a Novel Challenge}
% \title{Towards More Realistic DeepFakes: Semantic Mismatch as a Novel Challenge}
%\title{DeepFake on Realistic Scenarios: Using Real Data to Create DeepFakes}

%\title{Audio-Visual DeepFake Detection in More Realistic Scenarios}
%\title{Advancing Deepfake Detection: A Multi-Class Classification Approach Incorporating Audio-Visual Semantic Inconsistency} 
%\title{Audio-visual DeepFake Detection in More Realistic Scenarios: Semantic Mismatch as a Novel Challenge} 
%\title{Advancing Deepfake Detection: Incorporating Audio-Visual Semantic Consistency} 
%\title{An Advancement in Audio-Visual Deepfake Detection: A Multi-Class Approach}  
%\title{An Advanced Multi-Class Approach in Audio-Visual Deepfake Detection} 
%\title{DeepFake Detection on Realistic Scenarios: Using Real Data to Generate DeepFakes}

% \author{Deshmukh Sharayu Nilesh\\
% Universidade da Beira Interior\\
% Covilh\~{a}, Portugal\\
% {\tt\small d.sharyu.nilesh@ubi.pt}
% % For a paper whose authors are all at the same institution,
% % omit the following lines up until the closing ``}''.
% % Additional authors and addresses can be added with ``\and'',
% % just like the second author.
% % To save space, use either the email address or home page, not both
% \and
% Tiago Roxo\\
% Universidade da Beira Interior\\
% Covilh\~{a}, Portugal\\
% {\tt\small tiago.roxo@ubi.pt}
% \and
% Kailash A. Hambarde\\
% Universidade da Beira Interior\\
% Covilh\~{a}, Portugal\\
% {\tt\small kailash.hambarde@ubi.pt}
% \and
% Joana Cabral Costa\\
% Universidade da Beira Interior\\
% Covilh\~{a}, Portugal\\
% {\tt\small joana.cabral.costa@ubi.pt}
% }

\author{Sharayu Nilesh Deshmukh,~Kailash A. Hambarde,~Joana C. Costa,~Hugo Proença, and~Tiago Roxo\\
Instituto de Telecomunicações, Universidade da Beira Interior, Portugal\\
{\tt\small \{d.sharyu.nilesh, kailash.hambarde, joana.cabral.costa, hugomcp, tiago.roxo\}@ubi.pt}}

\maketitle
\thispagestyle{empty}

%%%%%%%%% ABSTRACT
\begin{abstract}
Current DeepFake detection scenarios are mostly binary, yet data manipulation can vary across audio, video, or both, whose variability is not captured in binary settings. Four-class audio-visual formulations address this by discriminating manipulation type, but introduce an unresolved problem: models may rely solely on data source integrity to detect DeepFakes without evaluating their semantic consistency. If the DeepFake origin is not in the data source but in its content, can semantic mismatch be assessed by the state-of-the-art? This paper proposes a new evaluation setup, extending the four-class formulation by explicitly modeling semantic-level inconsistency between authentic modalities with the introduction of a new class: Real Audio--Real Video with Semantic Mismatch (\textbf{RARV-SMM}). We assess the robustness of state-of-the-art models in this new realistic DeepFake setting, using the FakeAVCeleb dataset, highlighting the limitations of existing approaches when faced with semantic mismatch data. We further introduce three RARV-SMM variants that expose distinct architectural vulnerabilities as audio-visual divergence increases. We also propose a semantic reinforcement strategy that incorporates the semantic mismatch class and ImageBind embeddings to probe whether an explicit semantic coherence signal improves detection across architectures with different detection strategies, on FakeAVCeleb and LAV-DF, contributing toward more realistic DeepFake detectors. The source code available at  \url{https://github.com/sharayu-20/deepfake-semantic-mismatch}.
\end{abstract}

\begin{figure}[t]
    \centering
    \includegraphics[width=\linewidth]{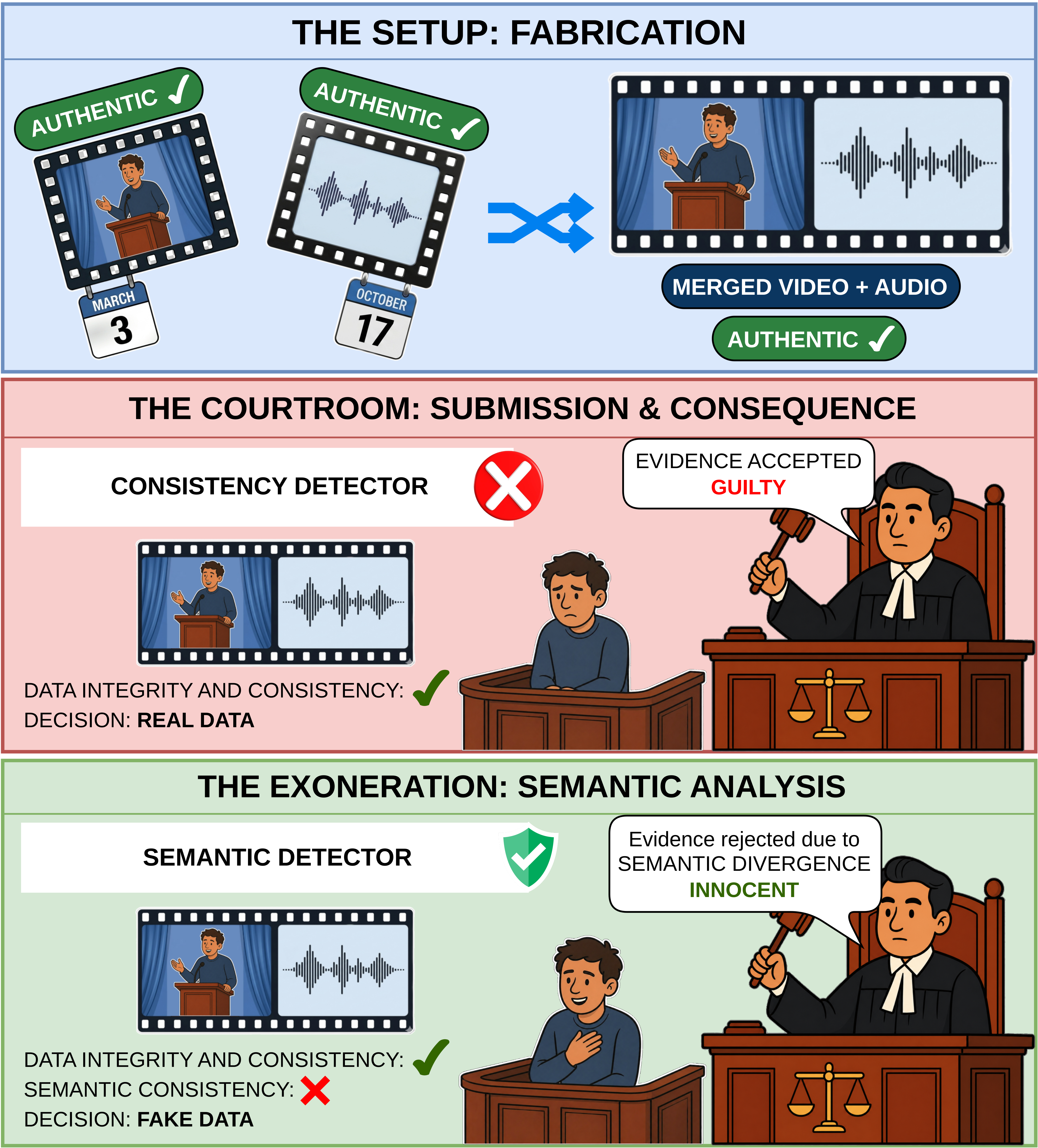}
    \caption{Illustration of the proposed threat: a four-class DeepFake detector inspects authentic video from one event and authentic audio from another, finds no synthesis artifacts in either stream, and classifies the 
    as safe, accepting potentially misleading content as genuine. This paper shows a vulnerability that semantic-aware detectors are better positioned to address, which is the basis of our proposal.}
    \label{fig:teaser}
\end{figure}

\section{Introduction}

Audio-visual content has become a critical medium for identity verification and political communication, raising concerns about the integrity of biometric systems that rely on facial/body appearance, voice, and lip dynamics~\cite{roxo2023exploring,roxo2024bias,roxo2025asdnb,MirskyLee,DeepfakeSurvey2,MultimodalTutorial}. The rapid proliferation of DeepFake technologies substantially elevates associated risks, modern systems synthesize realistic faces~\cite{FaceSwap,FaceReenact}, clone voices~\cite{VoiceClone}, and generate lip-synchronized forgeries~\cite{SyncNet}, enabling targeted disinformation, identity impersonation, and disbelief in audio-visual evidence~\cite{MirskyLee,MultimodalTutorial}. Detection methods have progressed from single modality artifact analysis~\cite{FaceForensics,LipsDoNotLie,RealForensics} toward four-class audio-visual formulations~\cite{FGMDF,AVDF,MRDF} that discriminate between Real Audio--Real Video (RARV), Real Audio--Fake Video (RAFV), Fake Audio--Real Video (FARV), and Fake Audio--Fake Video (FAFV), providing greater forensic interpretability than binary systems. However, all existing approaches share a fundamental unaddressed assumption, that all real audio--real video samples are semantically consistent.

This assumption fails in a class of realistic scenarios more dangerous than signal-level manipulations. We define semantic mismatch as the condition in which authentic audio and authentic video, each individually genuine, originate from different events, contexts, or narratives, such that their combined presentation conveys a false or misleading impression that neither source individually supports. %Unlike signal-level manipulation, semantic mismatch leaves no synthesis traces and cannot be identified by artifact-based detectors. 
A genuine speech recording of a public figure combined with genuine footage from a different event carries no synthesis artifacts and no signal-level manipulation; any audio-visual divergence~\cite{Agarwal2020, Bohacek2024} that arises is a natural consequence of combining authentic content from different contexts, rather than introduced through synthesis, yet it produces highly convincing misleading content that is entirely outside the scope of existing formulations~\cite{FGMDF,FGI,AVDF}. Our main motivation is illustrated in Figure~\ref{fig:teaser}, where an attacker can combine authentic video with authentic audio and both streams pass consistency inspection and the fabricated evidence is accepted as genuine, producing a false conclusion. With a semantic-aware detector, capable of reasoning over semantic divergence, it is possible to identify a new type of DeepFake data. Which raises the question: \textit{If the DeepFake origin is not in the data source but in its semantic content, are state-of-the-art models able to assess it?}

We propose a new evaluation setup, extending the four-class formulation by explicitly modeling semantic-level inconsistency between authentic modalities with the introduction a new class: Real Audio--Real Video with Semantic Mismatch (\textbf{RARV-SMM}). We evaluate FGMDF~\cite{FGMDF}, FGI~\cite{FGI}, and AVDF~\cite{AVDF} across different experimental settings, highlighting the limitations of existing approaches when faced with semantic mismatch data. Explicit five-class training improves performance in cross-modal and distance-based architectures, while data-source integrity-based methods lack the foundations needed to learn semantic reasoning. We also introduce three RARV-SMM variants of increasing audio-visual divergence, showing that semantic analysis alone is insufficient in more challenging configurations, and propose a semantic reinforcement strategy using frozen ImageBind~\cite{ImageBind} embeddings to probe the architecture-dependence of explicit semantic signals, finding consistent gains for graph-based and source-integrity models. The inclusion of the new class improves model performance in state-of-the-art settings, highlighting the importance of our contributions to DeepFake detection. 

We position RARV-SMM within the DeepFake detection framework because its threat model is identical to that of signal-level forgeries: fabricated audio-visual evidence intended to mislead. The manipulation here operates at the level of content curation rather than signal synthesis, yet the forensic question is whether audio and video originate from the same authentic event falls naturally within the scope of audio-visual integrity verification. Extending existing four-class formulations to account for this scenario is therefore a principled and necessary step, not a departure from the field. Our main contributions are three-fold:
\begin{itemize}
    \item We propose a novel audio-visual DeepFake detection framework that explicitly distinguishes semantic-level mismatch from signal-level manipulation, showing misclassification patterns that reveal the effectiveness of the underlying detection strategy of DeepFake detectors.

    \item We propose a semantic reinforcement strategy that incorporates the semantic mismatch class and augments the model's classifier with a frozen ImageBind cosine similarity score, introducing an explicit semantic coherence signal whose effectiveness is architecture-dependent. % : it improves robustness for graph-based and source-integrity architectures on the hardest variants and state-of-the-art settings, while revealing that distance-based architectures require a more tailored integration approach to benefit from such signals.

    \item We show that training in such new setting improves the performance of models with cross-modal approaches, while models that rely on source-data integrity cannot improve regardless of supervision, and the three considered variants expose distinct architectural vulnerabilities as audio-visual divergence increases.

\end{itemize}
 
%-------------------------------------------------------------------------
\section{Related Work}

\textbf{DeepFake Detection.} 
Early methods relied on unimodal analysis through CNNs and spatiotemporal networks trained on established benchmarks~\cite{FaceForensics,SpatioTemporal}, methods exploiting lip dynamics, irregular blinking, and physiological inconsistencies~\cite{Blinking,LipsDoNotLie,RealForensics}, and spectral approaches for synthetic speech detection~\cite{SonicSleuth}. Despite their contributions, these approaches cannot exploit cross-modal cues and are inherently limited to signal-level artifacts within a single modality. The growing sophistication of forgeries requiring simultaneous audio-video manipulation~\cite{MultimodalTutorial,MirskyLee} motivated the shift to joint audio-visual learning, where methods exploit synchronization patterns, cross-modal attention, and contrastive objectives~\cite{Zhou2021,MRDF,AVContrastive,VoiceFaceHomogeneity} to capture forgery traces that neither modality reveals alone. Ensemble-based approaches~\cite{MDS,AVDF} that independently aggregate unimodal predictions remain structurally limited, as they ignore complementary information from the relationship between streams. This leaves an entire class of realistic DeepFakes, where both streams are authentic but semantically inconsistent are entirely unaddressed, which is the basis of our proposal.

\textbf{Binary Approaches.}
In binary settings, models assess the synchronization between speech and facial motion, treating detection as a lip-sync consistency problem. Temporal audio-lip correspondence is exploited at multiple granularities, from clip level~\cite{LipSync} to frame-level phoneme-viseme alignment~\cite{Agarwal2020}. Zhou \etal~\cite{Zhou2021} model lip-speech alignment at multiple feature levels via contrastive learning; Bohacek \etal~\cite{Bohacek2024} measure mismatch by transcribing both audio and lip movements to text; and self-supervised methods pre-trained on large-scale authentic data~\cite{RealForensics,AVContrastive} improve generalization across unseen manipulations. Despite their contributions, binary approaches cannot distinguish which modality is manipulated and, more critically, implicitly assume all samples are consistent.

\begin{figure*}[t]
    \centering
    \includegraphics[width=\linewidth]{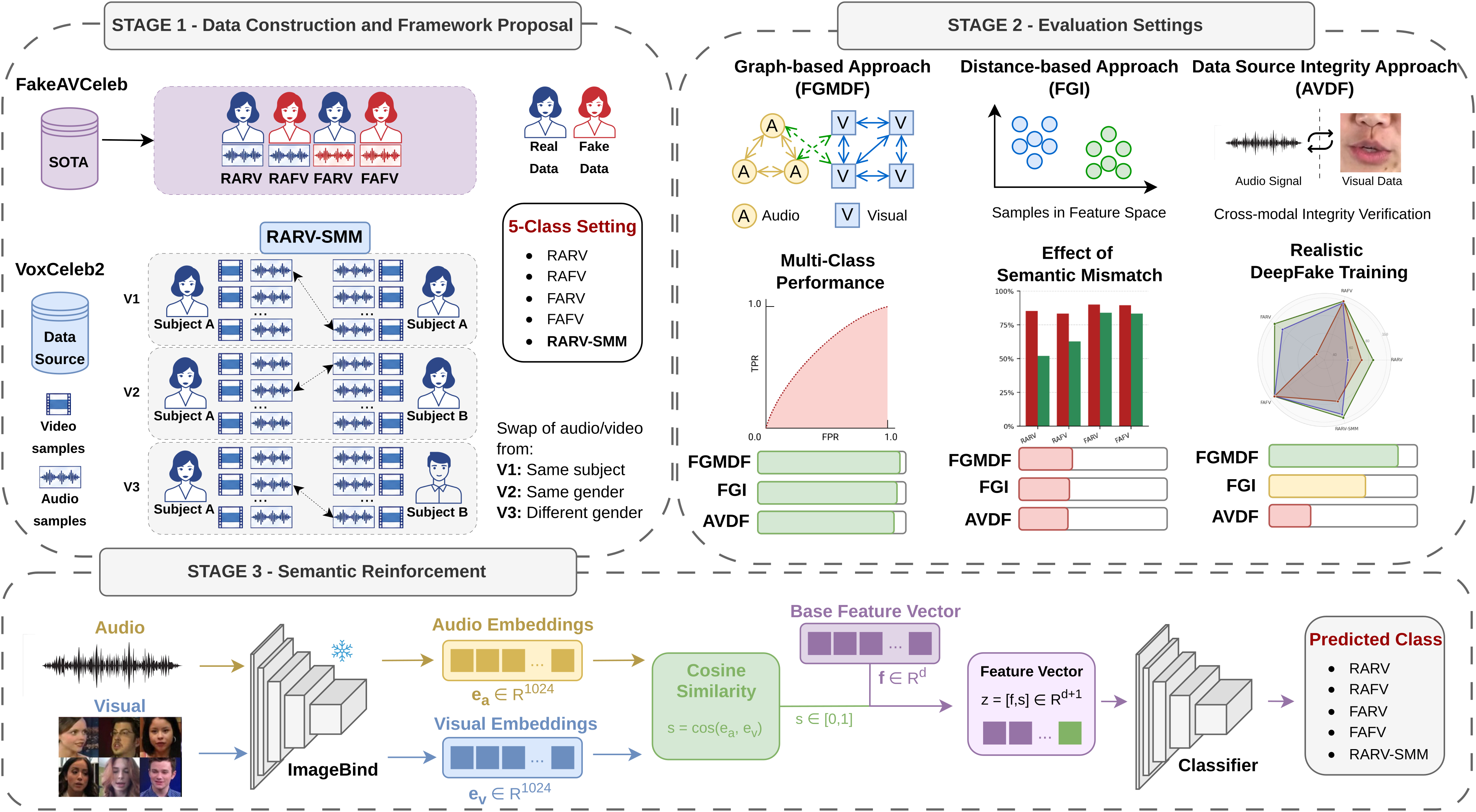}
    \caption{\textbf{Overview of the proposed five-class audio-visual DeepFake detection framework.} \textbf{Stage 1} constructs RARV-SMM samples from VoxCeleb2 across three variants (V1: same identity, different context; V2: different identity, same gender; V3: different identity, different gender) and integrates them with FakeAVCeleb to form a unified five-class dataset. \textbf{Stage 2} defines three experimental settings: multi-class performance, effect of introducing DeepFake class with semantic mismatch and realistic DeepFake challenge, and evaluates across three models with fundamentally different detection strategies. \textbf{Stage 3} applies a model-agnostic semantic reinforcement strategy using frozen ImageBind embeddings to augment each model's classifier.}
    \label{fig:overview}
\end{figure*}

\textbf{Multi-Class Approaches.} Here, audio, video, or both modalities can be manipulated. MRDF~\cite{MRDF} pioneering formalized this setting through cross-modality and within-modality regularization, yet can be seen as an extension of AVDF~\cite{AVDF} (using an AV-HuBERT~\cite{AVHuBERT} backbone grounding detection in lip-sync and source-data integrity signals). FGMDF~\cite{FGMDF} advances the paradigm with a heterogeneous graph-based architecture propagating information within and across audio visual nodes through dual cross-graph attention. FGI~\cite{FGI} computes spatially-local audio-visual discordance maps and was adapted in this work to the four-class formulation. FAUForensics~\cite{FAUForensics} and StatAVDF~\cite{StatAVDF} further extend four-class detection with physiological and statistical cues respectively. Despite their architectural diversity, all these formulations treat all real audio--real video samples as a single homogeneous class, with no mechanism to detect cases where authentic audio and authentic video originate from semantically inconsistent contexts and this is the gap our proposed RARV-SMM class directly addresses.
%-------------------------------------------------------------------------

\section{Proposed Approach}

The proposed empirical evaluation scenario, the experiments and the observed model improvements are summarized in Figure~\ref{fig:overview}. Stage~1 obtains the RARV-SMM class from VoxCeleb2 across three variants of increasing semantic divergence and integrates it with FakeAVCeleb to form the five class dataset. Stage~2 defines three experimental settings that progressively reveal how models respond to the new class by evaluating three models representing distinct detection philosophies. Finally, stage~3 introduces a model-agnostic semantic reinforcement strategy using frozen ImageBind embeddings to complement each model's classifier with an explicit semantic coherence signal.

\subsection{Problem Formulation}

Most DeepFake detection scenarios are binary, treating content as either fake or real. While four-class audio-visual formulations address the variability of manipulation type as discriminating audio manipulation, video manipulation, or both, they may motivate models to solely rely on data source integrity to detect DeepFakes, without evaluating the semantic consistency of the content. We propose a new evaluation setting by including a new fifth class, \textbf{RARV-SMM}, to probe whether state-of-the-art models can assess mismatches at a semantic level when both modalities are authentic, whose class definitions are summarized in Table~\ref{tab:formulation}.

Formally, class~5 (RARV-SMM) differs from class~1 (RARV) not at the signal level but semantically as both modalities are authentic yet originate from different contexts, events, or narratives. On RARV-SMM samples any low-level audio-visual divergence that arises, such as lip motion misalignment or acoustic differences between streams is an inherent and expected consequence of combining authentic content from semantically inconsistent sources, not the product of signal-level manipulation. The defining characteristic of RARV-SMM is that its inconsistency is rooted at the semantic level, requiring models to reason about cross-modal content rather than data-source manipulation traces.

\begin{table}[t]
\small
\centering
\caption{Characteristics of the proposed five-class audio-visual DeepFake formulation.}
\vspace{0.2cm}
\label{tab:formulation}
\begin{tabular}{@{}lccccc@{}}
\toprule
\multirow{2}{*}{Class} & \multicolumn{2}{c}{Real} & \multicolumn{2}{c}{Fake} & Semantic \\
 & Audio & Video & Audio & Video & Mismatch \\
\midrule
RARV     & \checkmark & \checkmark & $\times$ & $\times$ & $\times$ \\
RAFV     & \checkmark & $\times$   & $\times$ & \checkmark & $\times$ \\
FARV     & $\times$   & \checkmark & \checkmark & $\times$ & $\times$ \\
FAFV     & $\times$   & $\times$   & \checkmark & \checkmark & $\times$ \\
\midrule
RARV-SMM & \checkmark & \checkmark & $\times$ & $\times$ & \checkmark \\
\bottomrule
\end{tabular}
\end{table}

\begin{figure}[t]
    \centering
    \includegraphics[width=0.99\columnwidth]{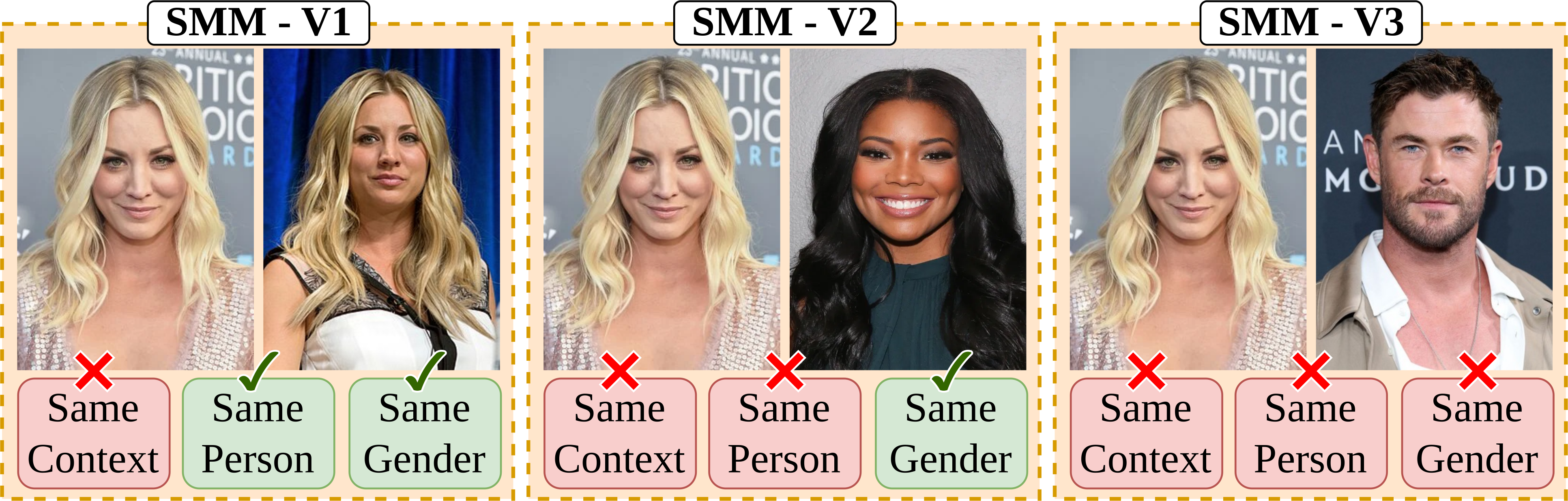}
    \caption{Illustration of the proposed RARV-SMM variants. \textbf{V1}: Audio and video belong to the same identity but different contexts. \textbf{V2}: Audio and video correspond to different identities of the same gender. \textbf{V3}: Audio and video correspond to identities of different gender.}
    \label{fig:rarv_smm_variants}
\end{figure}

\subsection{Proposed Semantic Class}
\label{subsec:semantic_classes}

\textbf{New Class Creation.} VoxCeleb2~\cite{VoxCeleb2} is the source for all RARV-SMM samples. It contains over 1 million utterances from 6,112 speakers collected in-the-wild from YouTube, with rich metadata on speaker identity, nationality, and recording context (red carpet interviews, stadium speeches, studio recordings) that is critical for our construction protocol. A total of 5,996 RARV-SMM samples are generated by first standardizing all clips. Video is resized to $224{\times}224$ pixels at 25 fps with durations of 3--10 seconds, and audio is converted to 16~kHz mono and loudness-normalized to $-23$~LUFS, with duration aligned to the corresponding video segment, then pairing authentic audio and video under identity and demographic constraints. Since the two streams may differ in duration, simple synchronization is applied that is, if audio exceeds video length it is truncated, and if audio is shorter it is looped or speed-adjusted using a minor \texttt{atempo} modification, maintaining the authentic, unmanipulated nature of both modalities while introducing semantic inconsistency between them. We note that pairing clips from different contexts naturally introduces some degree of low-level audio-visual divergence, such as lip motion misalignment or atempo-adjusted rhythm as an inherent consequence of the semantic mismatch. This is consistent with real-world scenarios where authentic content from different events is combined, and does not constitute signal-level manipulation.

\textbf{Class Variations.} To systematically analyze the challenge across increasing difficulty, we define three variants, as illustrated in Figure~\ref{fig:rarv_smm_variants}. \textbf{V1} pairs audio and video of the same person in different contexts (Event A \textit{vs.}\ Event B), forming the baseline configuration. \textbf{V2} pairs audio from one identity with video of a different identity of the same gender, increasing acoustic divergence. \textbf{V3} pairs audio and video from identities of different genders, maximizing the audio-visual divergence. These variants represent increasing degrees of audio-visual divergence between two authentic streams that are, in all cases, semantically mismatched. V1 introduces contextual semantic mismatch while preserving identity consistency, the purest instantiation of the proposed class. V2 and V3 extend this by adding identity and demographic divergence, which increases the overall cross-modal distance and introduces additional low-level cues (e.g., voice-face incompatibility) alongside the contextual mismatch. All three variants are instances of RARV-SMM; they do not represent different problem types but probe model sensitivity across a spectrum of audio-visual divergence.

\subsection{Selected State-of-the-Art Models}
\label{subsec:sota_models}

We evaluate three models that represent distinct detection philosophies, each making fundamentally different assumptions about what constitutes a DeepFake signal. \textbf{FGMDF}~\cite{FGMDF} formulates audio-visual DeepFake detection as a heterogeneous graph-based cross-modal reasoning problem. It represents audio and video as interconnected nodes and applies a graph attention network with intra-modal aggregation, cross-modal interaction, and graph pooling. This design explicitly models structural and synchronization relationships between modalities, enabling effective cross-modal inconsistency detection. \textbf{FGI}~\cite{FGI} originally proposed for binary detection, was adapted in this work to the four-class formulation to represent a distance-based detection strategy. FGI detects DeepFakes by measuring audio-visual discordance. FGI operates on whole-clip distance rather than structured cross-modal representations, it assesses audio-visual consistency through spatial proximity in feature space rather than semantic alignment. \textbf{AVDF}~\cite{AVDF} a self-supervised model pre-trained for audio-visual speech recognition, for DeepFake classification. Audio and video are each encoded by AV-HuBERT. Because AV-HuBERT was pre-trained to model lip-sync and speech correspondence, AVDF's detection is grounded in source-data integrity signals, specifically whether the audio and video streams are consistent with genuine lip-synced speech rather than higher-level semantic reasoning.

\subsection{Semantic Reinforcement}
\label{sec:method}

\textbf{Motivation.} The introduction of RARV-SMM creates a new challenge for models that rely on structural audio-visual correspondence alone may not fully distinguish semantically inconsistent authentic pairs from consistent ones, particularly as acoustic divergence increases across variants. We therefore propose a semantic reinforcement strategy to evaluate whether augmenting each model's classifier with an explicit semantic coherence signal improves five-class performance, and to characterise how this depends on each model's underlying detection mechanism.

\textbf{Approach.} The core observation is that existing model representations capture signal-level audio-visual correspondence but lack access to higher-level semantic alignment between what is said and what is shown. We address this by augmenting model's final classifier with a single additional feature that is a frozen ImageBind~\cite{ImageBind} cosine similarity score measuring semantic coherence between the audio and video streams. For each sample, a frozen pre-trained ImageBind model extracts a 1024-dimensional audio embedding $\mathbf{e}_a$, and a 1024-dimensional mean video frame embedding $\mathbf{e}_v$. Their cosine similarity is computed offline and normalized to $[0,1]$:

\begin{equation}
    s = \frac{1}{2}\left(
    \frac{\mathbf{e}_a \cdot \mathbf{e}_v}
    {\|\mathbf{e}_a\|\,\|\mathbf{e}_v\|} + 1
    \right) \in [0,\,1].
\end{equation}

% The raw cosine similarity from PyTorch's \texttt{F.cosine\_similarity} naturally lies in $[-1, 1]$, where $-1$ indicates completely opposite embeddings, $0$ indicates orthogonal (unrelated) embeddings, and $+1$ indicates perfectly matching embeddings. The $+1$ and $\frac{1}{2}$ terms remap this range to $[0, 1]$, providing a bounded, non-negative score where $0$ represents maximum semantic divergence, $0.5$ represents a neutral relationship, and $1$ represents full semantic alignment. This normalization ensures the semantic score is numerically compatible with the model's existing fusion representations that is liner classification setup.

The raw cosine similarity naturally lies in $[-1, 1]$. The $+1$ and $\frac{1}{2}$ terms remap this range to $[0, 1]$, providing a bounded, non-negative score where $0$ represents maximum semantic divergence, $0.5$ represents a neutral relationship, and $1$ represents full semantic alignment. This normalization ensures the semantic score is numerically compatible with the model's existing fusion representations that is linear classification setup. These scores are pre-computed, adding no overhead to training. At classification time, $s$ is concatenated with each model's fusion output $\mathbf{f}$, forming an enriched representation passed to a modified final classification layer. ImageBind is never fine-tuned and is not involved in backpropagation. The semantic score acts purely as an input feature, providing each model's classifier with an explicit semantic coherence signal that is entirely complementary to the structural correspondence already captured by the base architecture, enabling a direct evaluation of whether, and under what architectural conditions, an explicit semantic-level signal improves five-class detection.

\subsection{Implementation Details}

The five-class dataset comprises 21,373 samples across RARV, RAFV, FARV, FAFV, and RARV-SMM, following the same distribution as the original dataset. FGMDF~\cite{FGMDF} applies three Dual Cross-Graph Attention blocks over 10 video and 4 audio nodes, producing a 256-dimensional fusion vector to which the semantic score is appended, trained with Adam ($\text{lr}{=}1.1{\times}10^{-3}$), StepLR ($\gamma{=}0.5$, step${=}8$), and class-weighted cross entropy for 20 epochs at batch size 16. FGI~\cite{FGI} computes element-wise Euclidean distance over a $28{\times}28$ spatial grid to produce a 784-dimensional descriptor, appends the semantic score, and is trained with Adam ($\text{lr}{=}1{\times}10^{-4}$), and early stopping for 100 epochs at batch size 32. AVDF~\cite{AVDF} fine-tunes AV-HuBERT~\cite{AVHuBERT} with concatenated 768-dimensional audio-video embeddings passed through a Transformer encoder and a Linear layer, trained with Adam ($\text{lr}{=}2{\times}10^{-4}$), and early stopping for 30 epochs at batch size 64. For the semantic reinforcement, a frozen ImageBind~\cite{ImageBind} model generates semantic alignment scores as normalized cosine similarity between 1024-dimensional audio and video embeddings, mapped to $[0,1]$ and pre-computed offline for all samples.  All experiments used random seed 42 and were conducted on NVIDIA CUDA-enabled GPUs using PyTorch.

\section{Experiments}

\begin{table}[t]
\footnotesize
\centering
\caption{Performance of models on Multi-Class evaluation, considering only the 4-class scenario, on FakeAVCeleb dataset.}
\vspace{0.2cm}
\label{tab:baseline}
\begin{tabular}{@{}llccccc@{}}
\toprule
Model & Class & PR & RE & F1 & ACC & AUC \\
\midrule
\multirow{4}{*}{FGMDF}
 & RARV & 93.68 & 90.82 & 92.23 & \multirow{4}{*}{99.46} & \multirow{4}{*}{99.98} \\
 & RAFV & 100.00 & 96.97 & 98.46 & & \\
 & FARV & 99.52 & 99.68 & 99.60 & & \\
 & FAFV & 100.00 & 100.00 & 99.90 & & \\
\midrule
\multirow{4}{*}{FGI}
 & RARV & 64.63 & 76.81 & 70.19 & \multirow{4}{*}{95.15} & \multirow{4}{*}{98.98} \\
 & RAFV & 93.48 & 97.90 & 95.64 & & \\
 & FARV & 80.00 & 85.71 & 82.75 & & \\
 & FAFV & 99.30 & 94.01 & 96.58 & & \\
\midrule
\multirow{4}{*}{AVDF}
 & RARV & 81.22 & 89.75 & 85.27 & \multirow{4}{*}{87.00} & \multirow{4}{*}{97.47} \\
 & RAFV & 88.02 & 79.00 & 83.27 & & \\
 & FARV & 87.47 & 92.50 & 89.91 & & \\
 & FAFV & 92.29 & 86.75 & 89.43 & & \\
\bottomrule
\end{tabular}
\end{table}

\subsection{Datasets, Models, Settings, and Metrics}

\textbf{Datasets and Models.} We conduct experiments on two audio-visual DeepFake benchmarks and construct our RARV-SMM class from an authentic large-scale speaker dataset. \textbf{FakeAVCeleb}~\cite{FakeAVCeleb} is the primary benchmark for all main experiments. It contains 500 real videos and 19,500 fake videos from 500 celebrity identities across four ethnic backgrounds. Since FakeAVCeleb was constructed from VoxCeleb, it shares the same authentic source domain as our RARV-SMM samples, making it the natural benchmark to probe semantic mismatch robustness alongside signal-level manipulation detection. \textbf{LAV-DF}~\cite{LAVDF} is used to evaluate the contribution of the five-class formulation and semantic reinforcement against baselines. It contains 36,431 real and 99,873 fake clips with content-driven manipulations where fake segments constitute only 0.8--1.6 seconds on average. The models used in the experiments are described in section~\ref{subsec:sota_models}.

\textbf{Settings.} Three experimental settings are designed to progressively reveal how models respond to the new class: 1) Multi-Class Model Performance; 2) Effect of Semantic Mismatched Class; and 3) Realistic DeepFake Training. The first setting establishes the baseline by training and testing on the original four classes. The next setting evaluates four-class trained models directly on five-class test data without any retraining to assess how each model's internal representations map the new class to, revealing their underlying detection strategy. The final setting performs full five-class training and testing, assessing whether each model's architecture provides a sufficient pathway to learn semantic differences, which we also extend with RARV-SMM variants V1, V2, and V3.

\textbf{Metrics.} All models are evaluated using overall Accuracy (ACC), macro-averaged AUC (one-vs-rest), and per class Precision (PR), Recall (RE), and F1-Score (F1) to capture both global and class-specific performance under class-imbalanced conditions.

\subsection{Multi-Class Baseline Model Performance}

To have a baseline to compare the effect of the proposed evaluation setup, we first assess the performance of state-of-the-art models on the existing 4-class settings, in Table~\ref{tab:baseline}. All three models achieve strong performance under the standard four-class formulation, confirming the effectiveness of existing approaches. FGMDF achieves the highest accuracy and near-perfect AUC, reflecting its graph-based cross-modal representations. FGI and AVDF show comparatively lower RARV precision and recall, consistent with less explicit cross-modal semantic modeling. The results suggest that the four-class paradigm is well-solved under its own assumptions; the critical question is what happens when those assumptions are broken.

\begin{table}[t]
\setlength{\tabcolsep}{4pt}
\footnotesize
\centering
\caption{Overall ACC and AUC in 4-class (Multi-Class) and 5-class (Effect of SMM) scenarios on FakeAVCeleb dataset.}
\vspace{0.2cm}
\label{tab:4445_global}
\begin{tabular}{@{}lcccc@{}}
\toprule
\multirow{3}{*}{Model} & \multicolumn{2}{c}{ACC} & \multicolumn{2}{c}{AUC} \\
\cmidrule(l){2-3} \cmidrule(l){4-5}
 & Multi-Class & Effect of SMM & Multi-Class & Effect of SMM \\
\midrule
FGMDF & 99.46 & 80.07 & 99.98 & 95.55 \\
FGI   & 95.15 & 71.57 & 98.98 & 87.79 \\
AVDF  & 87.00 & 68.50 & 97.47 & 85.69 \\
\bottomrule
\end{tabular}
\end{table}

\begin{table}[t]
\footnotesize
\centering
\caption{Per-Class F1-Score in 4-class (Multi-Class) and 5-class (Effect of SMM) scenarios on FakeAVCeleb dataset. $\Delta$ represents the difference between Effect of SMM and Multi-Class.}
\vspace{0.2cm}
\label{tab:4445_perclass}
\begin{tabular}{@{}llccc@{}}
\toprule
Model & Class & Multi-Class & Effect of SMM & $\Delta$ \\
\midrule
\multirow{4}{*}{FGMDF}
 & RARV & 92.23 & 17.73 & $-$74.50 \\
 & RAFV & 98.47 & 99.49 & $+$01.02 \\
 & FARV & 99.60 & 98.72 & $-$00.88 \\
 & FAFV & 99.90 & 99.97 & $+$00.07 \\
\cmidrule(l){1-5}
\multirow{4}{*}{FGI}
 & RARV & 70.19 & 32.04 & $-$38.15 \\
 & RAFV & 95.64 & 74.04 & $-$21.60 \\
 & FARV & 82.75 & 81.08 & $-$01.67 \\
 & FAFV & 96.58 & 96.91 & $+$00.33 \\
\cmidrule(l){1-5}
\multirow{4}{*}{AVDF}
 & RARV & 85.27 & 60.12 & $-$25.15 \\
 & RAFV & 83.27 & 68.61 & $-$14.66 \\
 & FARV & 89.91 & 91.95 & $+$02.04 \\
 & FAFV & 89.43 & 91.74 & $+$02.31 \\
\bottomrule
\end{tabular}
\end{table}

\subsection{Semantic Mismatch on Baseline Models}

To assess how models perform with the data containing semantic mismatch, we introduce the RARV-SMM at test time without retraining and evaluate the extent how each model relies on data-source integrity in Table~\ref{tab:4445_global}. Although a decline in performance is to be expected, Table~\ref{tab:4445_perclass} goes further by identifying the classes with which new data is misclassified, revealing the underlying detection strategy employed by each model. For FGMDF, degradation is almost entirely concentrated in RARV, since RARV-SMM shares the same authentic data-source profile and the graph network assigns it to RARV accordingly. For FGI and AVDF, degradation spreads across both RARV and RAFV, suggesting these models place greater weight on the audio-visual origin relationship, such that mismatched audio perturbs this in ways resembling fake-video patterns.

% \begin{table}[t]
% \small
% \centering
% \caption{Per-Class F1-Score in 5-class scenario, considering a model with training on the proposed settings, on FakeAVCeleb dataset.}
% \label{tab:55_detail}
% \vspace{0.2cm}
% \begin{tabular}{@{}llc@{}}
% \toprule
% Model & Class & F1 (\%) \\
% \midrule
% \multirow{5}{*}{FGMDF}
%  & RARV     & 84.87 \\
%  & RAFV     & 99.49 \\
%  & FARV     & 99.29 \\
%  & FAFV     & 99.82 \\
%  & RARV-SMM & \textbf{98.34} \\
% \cmidrule(l){1-3}
% \multirow{5}{*}{FGI}
%  & RARV     & 22.50 \\
%  & RAFV     & 90.13 \\
%  & FARV     & 70.17 \\
%  & FAFV     & 97.33 \\
%  & RARV-SMM & \textbf{87.54} \\
% \cmidrule(l){1-3}
% \multirow{5}{*}{AVDF}
%  & RARV     & 52.02 \\
%  & RAFV     & 62.65 \\
%  & FARV     & 83.90 \\
%  & FAFV     & 83.23 \\
%  & RARV-SMM & \textbf{57.01} \\
% \bottomrule
% \end{tabular}
% \end{table}

\subsection{Realistic DeepFake Training}

\begin{table}[t]
\small
\centering
\caption{Overall ACC and AUC in 5-class scenario, considering a model without (Effect of SMM) and with (Real) training on our settings, on FakeAVCeleb dataset.}
\vspace{0.2cm}
\label{tab:55_global}
\begin{tabular}{@{}lcccc@{}}
\toprule
 & \multicolumn{2}{c}{ACC} & \multicolumn{2}{c}{AUC} \\
\cmidrule(l){2-3} \cmidrule(l){4-5}
Model & Effect of SMM & Real & Effect of SMM & Real \\
\midrule
FGMDF & 80.07 & 98.80 & 95.55 & 99.88 \\
FGI   & 71.57 & 91.18 & 87.79 & 98.15 \\
AVDF  & 68.50 & 67.65 & 85.69 & 90.44 \\
\bottomrule
\end{tabular}
\end{table}

\begin{table}[t]
\small
\centering
\caption{Per-Class F1-Score in 5-class scenario, considering a model with training on the proposed settings, on FakeAVCeleb dataset.}
\label{tab:55_detail}
\vspace{0.2cm}
\begin{tabular}{@{}lccc@{}}
    \toprule
    \multirow{3}{*}{Class} & \multicolumn{3}{c}{Model} \\
    \cmidrule(l){2-4}
    & FGMDF & FGI & AVDF \\
    \midrule
    RARV & 84.87 & 22.50 & 52.02 \\
    RAFV & 99.49 & 90.13 & 62.65 \\
    FARV & 99.90 & 70.17 & 83.90 \\
    FAFV & 99.82 & 97.33 & 83.23 \\
    RARV-SMM & 98.34 & 87.54 & 57.01 \\
    \bottomrule
\end{tabular}
\end{table}

\begin{figure}[t]
\centering
\includegraphics[width=\linewidth]{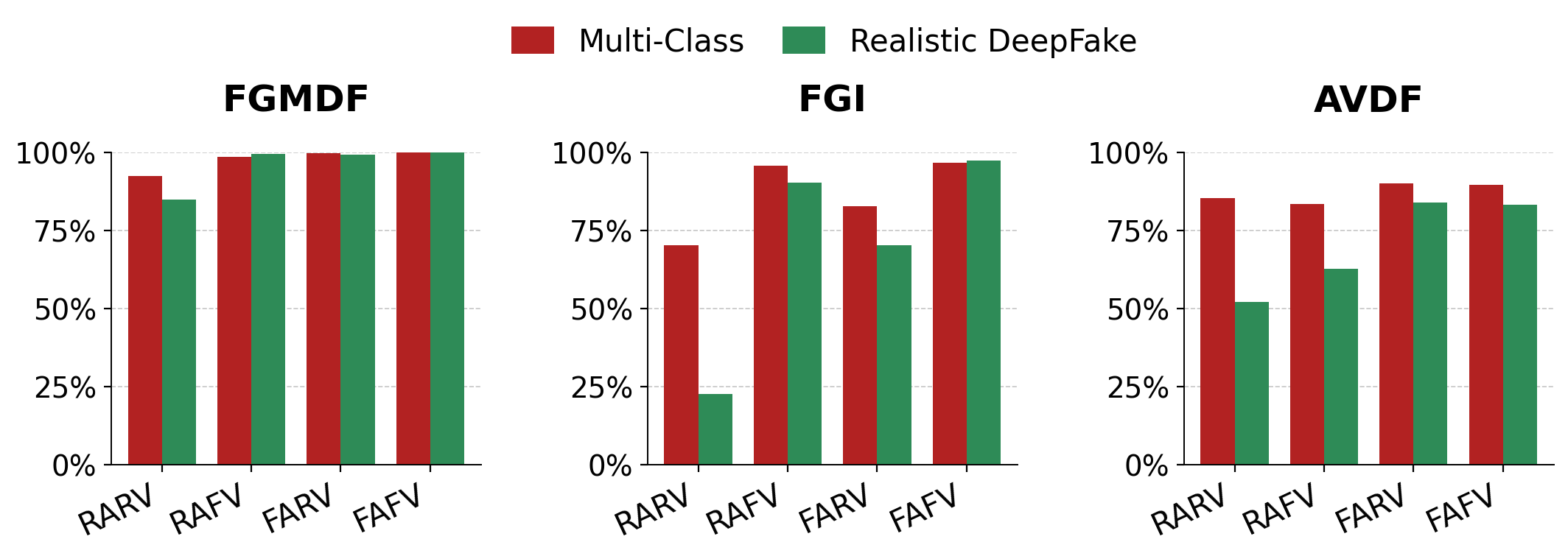}
\caption{Per-class F1 comparison between Multi-Class and Realistic DeepFake settings for FGMDF, FGI, and AVDF.}
\label{fig:f1_comparison}
\end{figure}

We extend our experiments to include the five-class training, in Tables~\ref{tab:55_global} and~\ref{tab:55_detail}, to evaluate if models can reliably capture the semantic differences of the proposed class. Improvement with explicit five-class supervision is expected for FGMDF and FGI, as their cross-modal and distance-based architectures provide a pathway to learn semantic distinctions. AVDF, however, cannot improve significantly, confirming that its lip-sync grounded representations rely on data-source integrity and cannot track semantic changes regardless of explicit supervision. FGMDF achieves the strongest RARV-SMM performance through its graph-based cross-modal reasoning, while FGI also learns the class effectively via distance-based analysis, though less robustly. The slight RARV drop for FGMDF is justified by class proximity as RARV-SMM and RARV share identical source-level profiles, keeping their graph representations close. FGI's greater RARV drop reflects some limitations of its distance-based approach, since in V1 settings the audio-video discordance for RARV-SMM is minimal (same identity), making it nearly indistinguishable from RARV and harder to correctly identify genuinely consistent pairs. Figure~\ref{fig:f1_comparison} confirms the per-class F1 shift from Multi-Class to Realistic DeepFake: FGMDF's drop concentrates exclusively in RARV, FGI's spreads across RARV and adjacent classes due to distance-based confusion, and AVDF's declines uniformly across all classes. The results suggest that each model captures semantic information differently, with some approaches proving more effective than others.

\subsection{Incremental Challenge of Semantic Variants}

\begin{figure*}[t]
\centering
\includegraphics[width=0.8\linewidth]{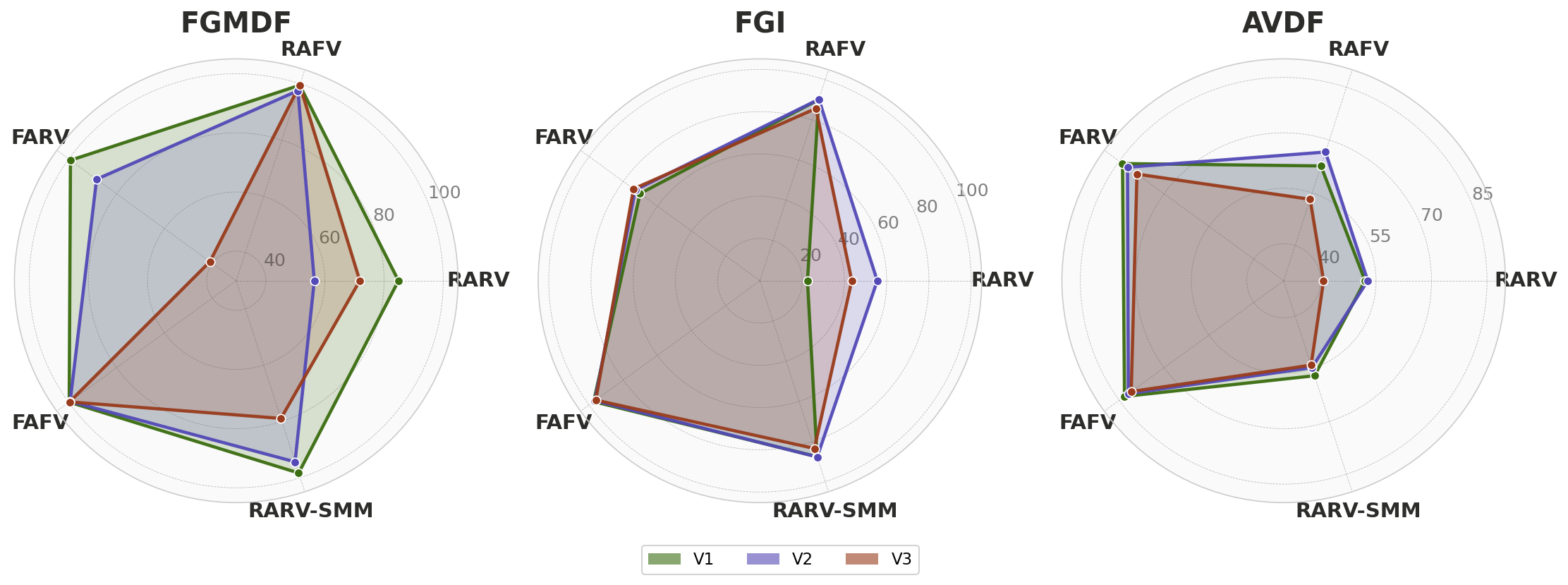}
\caption{Per-class F1 profiles across V1, V2 and V3 for FGMDF, FGI, and AVDF. Each axis represents one class; a larger polygon indicates better overall performance.}
\label{fig:radar_models}
\end{figure*}

To understand if the origin of the semantic mismatch influences models performance, we also extend our experiments, in Table~\ref{tab:v123_global}, towards the variations considered in section~\ref{subsec:semantic_classes}, semantic mismatch from the same person but in a different context (V1), same gender (V2), or different gender (V3). Moving from V1 to V2 and V3 increases the degree of audio-visual divergence, which affects each model differently depending on what signals underpin its detection strategy. Under V1, models can leverage the shared identity between audio and video to distinguish RARV-SMM from RARV. With V2 and V3, variations in the audio used to construct the fifth class increase its similarity to the other four classes, causing broader confusion that highlights the insufficiency of semantic analysis alone to distinguish between classes as audio variability increases. A full scope of evaluation is provided in Table~\ref{tab:v123_perclass}, where we show the mismatch between classes and the divergence of the variations. For FGMDF, same-gender voices in V2 raise confusion for RARV and FARV, and different-gender voices in V3 generalize this confusion most severely to FARV, where the voice-face mismatch resembles fake content. For FGI, greater acoustic distinctiveness paradoxically improves RARV clarity since larger voice differences reduce proximity between RARV-SMM and RARV in its distance-based space, though overall accuracy does not recover due to class imbalance. For AVDF, V3 raises broader confusion across all classes with emphasis on RARV and RAFV, as its lip-sync analysis cannot capture semantic differences and responds only to data-origin signals. Figure~\ref{fig:radar_models} confirms these patterns visually: for FGMDF, V1$\rightarrow$V2 shifts challenge to RARV and V1$\rightarrow$V3 collapses FARV most severely; for FGI, RARV clarity increases with more distinctive variants; for AVDF, V1$\rightarrow$V2 shows minimal change while V1$\rightarrow$V3 raises broader confusion on RARV and RAFV. 

\begin{table}[t]
\small
\centering
\caption{Overall ACC and AUC in the 5-class scenario across the three
variants, V1, V2, and V3, on the FakeAVCeleb dataset.}
\label{tab:v123_global}
\vspace{0.2cm}
\begin{tabular}{@{}lcccccc@{}}
\toprule
 & \multicolumn{2}{c}{V1} & \multicolumn{2}{c}{V2} & \multicolumn{2}{c}{V3} \\
\cmidrule(lr){2-3} \cmidrule(lr){4-5} \cmidrule(lr){6-7}
Model & ACC & AUC & ACC & AUC & ACC & AUC \\
\midrule
FGMDF & 98.80 & 99.88 & 96.73 & 99.26 & 93.58 & 99.51 \\
FGI   & 91.18 & 98.15 & 91.11 & 98.69 & 88.17 & 97.76 \\
AVDF  & 67.65 & 90.44 & 67.73 & 90.68 & 61.61 & 87.73 \\
\bottomrule
\end{tabular}
\end{table}

\begin{table}[t]
\setlength{\tabcolsep}{4pt}
\footnotesize
\centering
\caption{Per-class F1-Score in the 5-class scenario across the three variants, V1, V2, and V3, on the FakeAVCeleb dataset.}
\label{tab:v123_perclass}
\vspace{0.2cm}
\begin{tabular}{@{}llccccc@{}}
\toprule
Model & Class & V1 & V2 & V3 & V1$-$V2 & V1$-$V3 \\
\midrule
\multirow{5}{*}{FGMDF}
 & RARV     & 84.87 & 56.41 & 71.79 & $-$28.46 & $-$13.08 \\
 & RAFV     & 99.49 & 97.54 & 99.52 & $-$01.95  & $+$00.03  \\
 & FARV     & 99.29 & 88.37 & 40.99 & $-$10.92 & $-$58.30 \\
 & FAFV     & 99.82 & 99.42 & 99.67 & $-$00.40  & $-$00.15  \\
 & RARV-SMM & 98.34 & 94.38 & 78.94 & $-$03.96  & $-$19.40 \\
\cmidrule(l){1-7}
\multirow{5}{*}{FGI}
 & RARV     & 22.50 & 55.55 & 43.47 & $+$33.05 & $+$20.97 \\
 & RAFV     & 90.13 & 90.28 & 85.71 & $+$00.15  & $-$04.42  \\
 & FARV     & 70.17 & 73.01 & 73.97 & $+$02.84  & $+$03.80  \\
 & FAFV     & 97.33 & 96.61 & 96.10 & $-$00.72  & $-$01.23  \\
 & RARV-SMM & 87.54 & 87.68 & 83.46 & $+$00.14  & $-$04.08  \\
\cmidrule(l){1-7}
\multirow{5}{*}{AVDF}
 & RARV     & 52.02 & 52.75 & 40.80 & $+$00.73  & $-$11.22 \\
 & RAFV     & 62.65 & 66.60 & 53.16 & $+$03.95  & $-$09.49  \\
 & FARV     & 83.90 & 82.22 & 79.07 & $-$01.68  & $-$04.83  \\
 & FAFV     & 83.23 & 81.96 & 80.98 & $-$01.27  & $-$02.25  \\
 & RARV-SMM & 57.01 & 54.72 & 54.01 & $-$02.29  & $-$03.00  \\
\bottomrule
\end{tabular}
\end{table}

\subsection{Semantic Reinforcement Improvement}
Given that the lack of semantic analysis is the underlying issue for all models, we explore a semantic reinforcement approach to assess whether this complement increases model performance. Table~\ref{tab:sem_reinf} reports ACC and AUC across variants for models augmented with the semantic reinforcement strategy. The results reveal that the effectiveness of the ImageBind cosine similarity score is architecture-dependent: while the approach provides a viable semantic signal, its impact on classification is conditioned on how each model's decision space integrates an additional scalar feature. FGMDF, which already achieves strong performance, shows marginal but consistent gains across all variants, confirming that even a well-performing architecture benefits from an explicit semantic coherence signal.
%FGMDF, which already achieves strong performance, shows gains in V2 and V3, where higher audio-visual divergence gives the ImageBind similarity score more discriminative power. 
% In V1, where audio and video share the same identity and the mismatch is purely contextual, the semantic reinforcement yields a marginal difference within single-seed variance, consistent with V1 being the configuration in which the base architecture is already most capable}.
For FGI, semantic reinforcement leads to a substantial performance decrease across all variants. Although ImageBind provides a semantically meaningful signal, appending a cosine similarity scalar to FGI's feature vector is not a suitable integration approach for this architecture. FGI operates by measuring spatial proximity between audio and visual representations in feature space, its detection is inherently distance-based. Introducing an additional distance-like value into this space does not complement its reasoning; instead, it disrupts the learned distance geometry on which classification depends.
% This indicates that improving semantic reasoning in distance-based models requires a more tailored integration strategy beyond simple feature concatenation, and is identified as a direction for future work.} 
The most substantial gains are observed for AVDF, which achieves consistent ACC and AUC across all variants, demonstrating that the semantic complement directly compensates for the AV-HuBERT backbone's inability to capture semantic divergence through its lip-sync grounded representations. Together, these results show that the proposed semantic reinforcement is effective for graph-based and source-integrity architectures, while distance-based models require architecture-aware integration strategies to benefit from explicit semantic signals.

\begin{table}[t]
\small
\centering
\caption{Overall ACC and AUC using the proposed semantic reinforcement strategy, across the three variations, on FakeAVCeleb dataset.}
\label{tab:sem_reinf}
\vspace{0.2cm}
\begin{tabular}{@{}lcccccc@{}}
\toprule
 & \multicolumn{2}{c}{V1} & \multicolumn{2}{c}{V2} & \multicolumn{2}{c}{V3} \\
\cmidrule(lr){2-3} \cmidrule(lr){4-5} \cmidrule(lr){6-7}
Model & ACC & AUC & ACC & AUC & ACC & AUC \\
\midrule
FGMDF & 98.30 & 99.77 & 96.94 & 99.31 & 98.93 & 99.77 \\
FGI   & 74.34 & 92.61 & 74.81 & 92.52 & 77.52 & 93.97    \\
AVDF  & 73.47 & 93.46 & 73.06 & 93.50 & 73.00 & 93.37 \\
\bottomrule
\end{tabular}
\end{table}

\subsection{Comparison with State-of-the-Art}

To further extend the importance and relevance of the proposed class, we explore the influence of training in these new data towards the original 4-class settings. Beyond its forensic interpretability, the proposed five-class formulation with semantic reinforcement also yields direct performance improvements over existing multimodal DeepFake detection methods, as shown in Table~\ref{tab:sota}. Our approach matches state-of-the-art performance on FakeAVCeleb and substantially outperforms all baselines on LAV-DF, confirming two complementary contributions: 1) introducing RARV-SMM forces the model to develop more discriminative representations of genuine content, maintaining four-class detection robustness while learning a more complete threat model; and 2) the semantic reinforcement strategy generalizes beyond FakeAVCeleb to LAV-DF without any cross-dataset training, validating that learning to reason about semantic coherence is beneficial across datasets with fundamentally different construction methodologies and manipulation characteristics.

\begin{table}[t]
\small
\centering
\caption{Comparison with state-of-the-art multimodal methods (video-level ACC and AUC) under four-class evaluation on FakeAVCeleb and LAV-DF. Our model is FGMDF trained under the five-class setting and with semantic reinforcement. Bold and underline refer to the best and second best results, respectively.}
\label{tab:sota}
\vspace{0.2cm}
\begin{tabular}{@{}lcccc@{}}
\toprule
\multirow{3}{*}{Method} & \multicolumn{2}{c}{FakeAV} & \multicolumn{2}{c}{LAV-DF} \\
\cmidrule(lr){2-3} \cmidrule(lr){4-5}
 & ACC & AUC & ACC & AUC \\
\midrule
MDS~\cite{MDS}           & 93.71 & 73.18 & 65.10 & 80.91 \\
JAVDD~\cite{Zhou2021}    & 94.28 & 77.78 & 65.85 & 82.71 \\
BA-TFD~\cite{LAVDF}      & 93.77 & 77.57 & 66.62 & 86.52 \\
FGMDF~\cite{FGMDF}       & \textbf{99.91} & \textbf{99.97} & 72.13 & 89.56 \\
\midrule
% Ours (5-class) & \textbf{99.58} & \textbf{99.96} & \textbf{99.82} & 
% \textbf{99.13} \\
Ours (5-class) & 98.61 & 99.94 & \underline{92.80} & 
\textbf{99.15} \\
Ours (5-class + Semantic) & \underline{99.58} & \underline{99.96} & \textbf{92.82} & 
\underline{99.13} \\
\bottomrule
\end{tabular}
\end{table}

\section{Conclusion and Future Work}

This paper introduces a new audio-visual DeepFake detection formulation that explicitly models semantic mismatch between authentic audio and authentic video (RARV-SMM), probing whether state-of-the-art models rely entirely on data source integrity or can assess content-level inconsistencies. Through systematic evaluation of FGMDF, FGI, and AVDF across different settings, we established that state-of-the-art models fail on RARV-SMM, with misclassification patterns that diagnostically reveal each model's detection strategy. Explicit five-class training improves performance of models with cross-modal and distance-based architectures, where FGMDF and FGI learn semantic distinctions, while lip-sync strategies like AVDF cannot improve significantly. Three RARV-SMM variants confirm that semantic analysis alone is insufficient in more challenging configurations, and a semantic reinforcement strategy reveals an architecture-dependent relationship with semantic signals: graph-based and source-integrity architectures benefit from the explicit ImageBind cosine similarity feature, while distance-based architectures are hindered by it. % — as introducing an additional distance-like scalar disrupts the learned feature geometry on which their classification depends. This finding highlights that the integration of semantic signals into DeepFake detectors must be tailored to the underlying detection mechanism.} 
These findings establish semantic mismatch as an important and currently underexplored vulnerability in state-of-the-art audio-visual DeepFake detection systems, and demonstrate that explicitly modelling it as a distinct class improves detection robustness without sacrificing performance on standard benchmarks. As future work, we suggest extending this framework to more datasets, incorporate the five-class formulation into binary deployed systems, developing architecture-aware semantic integration strategies for distance-based models, and develop architectures with explicit semantic reasoning modules capable of representing narrative and identity-level coherence beyond temporal synchronization.

\section*{Acknowledgments}

This work is funded by national funds through FCT – Fundação para a Ciência e a Tecnologia, I.P., and, when eligible, co-funded by EU funds under project/support UID/50008/2025 – Instituto de Telecomunicações, with DOI identifier $<$https://doi.org/10.54499/UID/50008/2025$>$; and by the FCT Doctoral Grant 2021.04905.BD.

{\small
\bibliographystyle{ieee}
\bibliography{egbib}

@inproceedings{Agarwal2020,
  title={Detecting deep-fake videos from phoneme-viseme mismatches},
  author={Agarwal, Shruti and Farid, Hany and Fried, Ohad and Agrawala, Maneesh},
  booktitle={Proceedings of the IEEE/CVF conference on computer vision and pattern recognition workshops},
  pages={660--661},
  year={2020}
}

@misc{SonicSleuth,
  title={Audio deep fake detection with Sonic Sleuth model. Computers, 13 (10), 256},
  author={Alshehri, A and Almalki, D and Alharbi, E and Albaradei, S},
  journal={MDPI},
  year={2024}
}

@article{FGI,
  title={Detecting audio-visual deepfakes with fine-grained inconsistencies},
  author={Astrid, Marcella and Ghorbel, Enjie and Aouada, Djamila},
  journal={arXiv preprint arXiv:2408.06753},
  year={2024}
}

@inproceedings{StatAVDF,
  title={Statistics-aware audio-visual deepfake detector},
  author={Astrid, Marcella and Ghorbel, Enjie and Aouada, Djamila},
  booktitle={2024 IEEE International Conference on Image Processing (ICIP)},
  pages={2557--2563},
  year={2024},
  organization={IEEE}
}

@inproceedings{Bohacek2024,
  title={Lost in translation: Lip-sync deepfake detection from audio-video mismatch},
  author={Bohacek, Matyas and Farid, Hany},
  booktitle={Proceedings of the IEEE/CVF Conference on Computer Vision and Pattern Recognition},
  pages={4315--4323},
  year={2024}
}

@article{LAVDF,
  title={Glitch in the matrix: A large scale benchmark for content driven audio--visual forgery detection and localization},
  author={Cai, Zhixi and Ghosh, Shreya and Dhall, Abhinav and Gedeon, Tom and Stefanov, Kalin and Hayat, Munawar},
  journal={Computer Vision and Image Understanding},
  volume={236},
  pages={103818},
  year={2023},
  publisher={Elsevier}
}

@article{VoiceFaceHomogeneity,
  title={Voice-face homogeneity tells deepfake},
  author={Cheng, Harry and Guo, Yangyang and Wang, Tianyi and Li, Qi and Chang, Xiaojun and Nie, Liqiang},
  journal={ACM Transactions on Multimedia Computing, Communications and Applications},
  volume={20},
  number={3},
  pages={1--22},
  year={2023},
  publisher={ACM New York, NY}
}

@inproceedings{MDS,
  title={Not made for each other-audio-visual dissonance-based deepfake detection and localization},
  author={Chugh, Komal and Gupta, Parul and Dhall, Abhinav and Subramanian, Ramanathan},
  booktitle={Proceedings of the 28th ACM international conference on multimedia},
  pages={439--447},
  year={2020}
}

@article{VoxCeleb2,
  title={Voxceleb2: Deep speaker recognition},
  author={Chung, Joon Son and Nagrani, Arsha and Zisserman, Andrew},
  journal={arXiv preprint arXiv:1806.05622},
  year={2018}
}

@inproceedings{MultimodalTutorial,
  title={Multimodal Deepfake Generation and Detection: Challenges, Methods, and Future Directions},
  author={Dhall, Abhinav and Cai, Zhixi and Ghosh, Shreya},
  booktitle={Companion Proceedings of the 27th International Conference on Multimodal Interaction},
  pages={65--66},
  year={2025}
}

@inproceedings{AVContrastive,
  title={Self-supervised video forensics by audio-visual anomaly detection},
  author={Feng, Chao and Chen, Ziyang and Owens, Andrew},
  booktitle={proceedings of the IEEE/CVF conference on computer vision and pattern recognition},
  pages={10491--10503},
  year={2023}
}

@inproceedings{ImageBind,
  title={Imagebind: One embedding space to bind them all},
  author={Girdhar, Rohit and El-Nouby, Alaaeldin and Liu, Zhuang and Singh, Mannat and Alwala, Kalyan Vasudev and Joulin, Armand and Misra, Ishan},
  booktitle={Proceedings of the IEEE/CVF conference on computer vision and pattern recognition},
  pages={15180--15190},
  year={2023}
}

@inproceedings{SpatioTemporal,
  title={Spatiotemporal inconsistency learning for deepfake video detection},
  author={Gu, Zhihao and Chen, Yang and Yao, Taiping and Ding, Shouhong and Li, Jilin and Huang, Feiyue and Ma, Lizhuang},
  booktitle={Proceedings of the 29th ACM international conference on multimedia},
  pages={3473--3481},
  year={2021}
}

@inproceedings{RealForensics,
  title={Leveraging real talking faces via self-supervision for robust forgery detection},
  author={Haliassos, Alexandros and Mira, Rodrigo and Petridis, Stavros and Pantic, Maja},
  booktitle={Proceedings of the IEEE/CVF conference on computer vision and pattern recognition},
  pages={14950--14962},
  year={2022}
}

@inproceedings{LipsDoNotLie,
  title={Lips don't lie: A generalisable and robust approach to face forgery detection},
  author={Haliassos, Alexandros and Vougioukas, Konstantinos and Petridis, Stavros and Pantic, Maja},
  booktitle={Proceedings of the IEEE/CVF conference on computer vision and pattern recognition},
  pages={5039--5049},
  year={2021}
}

@article{VoiceClone,
  title={Transfer learning from speaker verification to multispeaker text-to-speech synthesis},
  author={Jia, Ye and Zhang, Yu and Weiss, Ron and Wang, Quan and Shen, Jonathan and Ren, Fei and Nguyen, Patrick and Pang, Ruoming and Lopez Moreno, Ignacio and Wu, Yonghui and others},
  journal={Advances in neural information processing systems},
  volume={31},
  year={2018}
}

@inproceedings{AVDF,
  title={Evaluation of an audio-video multimodal deepfake dataset using unimodal and multimodal detectors},
  author={Khalid, Hasam and Kim, Minha and Tariq, Shahroz and Woo, Simon S},
  booktitle={Proceedings of the 1st workshop on synthetic multimedia-audiovisual deepfake generation and detection},
  pages={7--15},
  year={2021}
}

@article{FakeAVCeleb,
  title={FakeAVCeleb: A novel audio-video multimodal deepfake dataset},
  author={Khalid, Hasam and Tariq, Shahroz and Kim, Minha and Woo, Simon S},
  journal={arXiv preprint arXiv:2108.05080},
  year={2021}
}

@inproceedings{Blinking,
  title={In ictu oculi: Exposing ai created fake videos by detecting eye blinking},
  author={Li, Yuezun and Chang, Ming-Ching and Lyu, Siwei},
  booktitle={2018 IEEE International workshop on information forensics and security (WIFS)},
  pages={1--7},
  year={2018},
  organization={Ieee}
}

@article{DeepfakeSurvey2,
  title={Evolving from single-modal to multi-modal facial deepfake detection: Progress and challenges},
  author={Liu, Ping and Tao, Qiqi and Zhou, Joey Tianyi},
  journal={arXiv preprint arXiv:2406.06965},
  year={2024}
}

@article{MirskyLee,
  title={The creation and detection of deepfakes: A survey},
  author={Mirsky, Yisroel and Lee, Wenke},
  journal={ACM computing surveys (CSUR)},
  volume={54},
  number={1},
  pages={1--41},
  year={2021},
  publisher={ACM New York, NY, USA}
}

@inproceedings{FaceSwap,
  title={Fsgan: Subject agnostic face swapping and reenactment},
  author={Nirkin, Yuval and Keller, Yosi and Hassner, Tal},
  booktitle={Proceedings of the IEEE/CVF international conference on computer vision},
  pages={7184--7193},
  year={2019}
}

@inproceedings{SyncNet,
  title={A lip sync expert is all you need for speech to lip generation in the wild},
  author={Prajwal, KR and Mukhopadhyay, Rudrabha and Namboodiri, Vinay P and Jawahar, CV},
  booktitle={Proceedings of the 28th ACM international conference on multimedia},
  pages={484--492},
  year={2020}
}

@inproceedings{FaceForensics,
  title={Faceforensics++: Learning to detect manipulated facial images},
  author={Rossler, Andreas and Cozzolino, Davide and Verdoliva, Luisa and Riess, Christian and Thies, Justus and Nie{\ss}ner, Matthias},
  booktitle={Proceedings of the IEEE/CVF international conference on computer vision},
  pages={1--11},
  year={2019}
}

@article{AVHuBERT,
  title={Learning audio-visual speech representation by masked multimodal cluster prediction},
  author={Shi, Bowen and Hsu, Wei-Ning and Lakhotia, Kushal and Mohamed, Abdelrahman},
  journal={arXiv preprint arXiv:2201.02184},
  year={2022}
}

@inproceedings{FaceReenact,
  title={Face2face: Real-time face capture and reenactment of rgb videos},
  author={Thies, Justus and Zollhofer, Michael and Stamminger, Marc and Theobalt, Christian and Nie{\ss}ner, Matthias},
  booktitle={Proceedings of the IEEE conference on computer vision and pattern recognition},
  pages={2387--2395},
  year={2016}
}

@article{FAUForensics,
  title={Fauforensics: Boosting audio-visual deepfake detection with facial action units},
  author={Wang, Jian and Wu, Baoyuan and Liu, Li and Liu, Qingshan},
  journal={IEEE Transactions on Information Forensics and Security},
  year={2026},
  publisher={IEEE}
}

@article{FGMDF,
  title={Fine-grained multimodal deepfake classification via heterogeneous graphs},
  author={Yin, Qilin and Lu, Wei and Cao, Xiaochun and Luo, Xiangyang and Zhou, Yicong and Huang, Jiwu},
  journal={International Journal of Computer Vision},
  volume={132},
  number={11},
  pages={5255--5269},
  year={2024},
  publisher={Springer}
}

@inproceedings{Zhou2021,
  title={Joint audio-visual deepfake detection},
  author={Zhou, Yipin and Lim, Ser-Nam},
  booktitle={Proceedings of the IEEE/CVF international conference on computer vision},
  pages={14800--14809},
  year={2021}
}

@inproceedings{MRDF,
  title={Cross-modality and within-modality regularization for audio-visual deepfake detection},
  author={Zou, Heqing and Shen, Meng and Hu, Yuchen and Chen, Chen and Chng, Eng Siong and Rajan, Deepu},
  booktitle={ICASSP 2024-2024 IEEE International Conference on Acoustics, Speech and Signal Processing (ICASSP)},
  pages={4900--4904},
  year={2024},
  organization={IEEE}
}

@inproceedings{LipSync,
  title={Out of time: automated lip sync in the wild},
  author={Chung, Joon Son and Zisserman, Andrew},
  booktitle={Asian conference on computer vision},
  pages={251--263},
  year={2016},
  organization={Springer}
}

@inproceedings{roxo2023exploring,
  title={On exploring audio anomaly in speech},
  author={Roxo, Tiago and Costa, Joana Cabral and In{\'a}cio, Pedro RM and Proen{\c{c}}a, Hugo},
  booktitle={2023 IEEE International Workshop on Information Forensics and Security (WIFS)},
  pages={1--6},
  year={2023},
  organization={IEEE}
}

@article{roxo2024bias,
  title={BIAS: A Body-Based Interpretable Active Speaker Approach},
  author={Roxo, Tiago and Costa, Joana C and In{\'a}cio, Pedro RM and Proen{\c{c}}a, Hugo},
  journal={IEEE Transactions on Biometrics, Behavior, and Identity Science},
  year={2024},
  publisher={IEEE}
}

@inproceedings{roxo2025asdnb,
  title={Asdnb: Merging face with body cues for robust active speaker detection},
  author={Roxo, Tiago and Costa, Joana C and In{\'a}cio, Pedro RM and Proen{\c{c}}a, Hugo},
  booktitle={2025 IEEE International Joint Conference on Biometrics (IJCB)},
  pages={1--10},
  year={2025},
  organization={IEEE}
}
}

\end{document}